# Learning the PE Header, Malware Detection with Minimal Domain Knowledge

Edward Raff · Jared Sylvester · Charles Nicholas

**Abstract** Many efforts have been made to use various forms of domain knowledge in malware detection. Currently there exist two common approaches to malware detection without domain knowledge, namely byte n-grams and strings. In this work we explore the feasibility of applying neural networks to malware detection and feature learning. We do this by restricting ourselves to a minimal amount of domain knowledge in order to extract a portion of the Portable Executable (PE) header. By doing this we show that neural networks can learn from raw bytes without explicit feature construction, and perform even better than a domain knowledge approach that parses the PE header into explicit features.

## 1 Introduction

It is often desirable to use domain knowledge when building a machine learning model for a classification task. Domain knowledge is any kind of information used, either in feature selection, processing, design choices, code written or libraries used, that is specific to the given problem domain. Such knowledge cannot often be applied to new domains, but using domain knowledge allows us to embed our prior beliefs into the system. This may aid our intuition as to what classification models will perform best, and what level of performance we might be able to expect. However, it can be difficult to effectively apply domain knowledge to the task of malware detection, which we define as building a classifier that indicates whether an executable binary is benign or malicious. The format and nature of executable files is complicated and not always consistent.[1] Just parsing a Portable Executable (PE) file correctly is difficult. While libraries exist to do this, their development is non-trivial and time intensive. Furthermore, the Windows operating system does not always enforce its own specification [20, 33], and that specification may be changed in the future, requiring additional work to update a system. These problems are only compounded by the fact that malware may intentionally violate the PE standard or Windows operating system (OS) loading process.

Given these difficulties, we seek to examine the practicability of the malware detection problem with minimal domain knowledge. Indeed, byte n-gramming is a technique commonly used in malware classifiers, in part because it requires no domain knowledge [31]. However, it has been shown that whole file n-gramming presents a computational burden and does not perform as well as previous results suggested [45].

In this work we evaluate the feasibility of learning with a minimal amount of domain knowledge using

E. Raff · J. Sylvester
Laboratory for Physical Sciences, 8050 Greenmead Drive College Park, MD 20740
E-mail: edraff@lps.umd.edu, E-mail: jared@lps.umd.edu

E. Raff · C. Nicholas
Computer Science and Electrical Engineering, University of Maryland, Baltimore County, 1000 Hilltop Circle Baltimore, MD 21250 USA
E-mail: raff.edward@umbc.edu, E-mail: nicholas@umbc.edu

E. Raff · J. Sylvester
Booz Allen Hamilton, Strategic Innovation Group
E-mail: raff_edward@bah.com
E-mail: sylvester_jared@bah.com

---

[1] Several types of files are "executable" under Windows. We abuse the terminology slightly in that when we refer to "executables" we mean executable binaries, the names of which normally have the "exe" or "dll" suffixes.

neural networks on raw byte sequences.[2] To do so, we restrict ourselves to only the PE header of an executable. Specifically, we consider only the *MS-DOS*, *COFF*, and *Optional* headers of the PE file. (Despite its name, the Optional header is usually required.) A review of other approaches using only header information or minimal domain knowledge for malware detection is given in section 2. In section 3 we describe our two baseline approaches, and the minimal amount of domain knowledge needed for our approaches. We then describe the specifics of our network architecture designs in section 4, which use only the raw bytes extracted from the PE header. The performance of all of these approaches is evaluated and discussed in section 5, where we show that our neural network achieves higher accuracies than our baselines without needing to use domain knowledge. To give further confidence to these results we exploit our architecture design in section 6 to provide evidence that our neural networks are learning concepts similar to those learned in the domain knowledge approach, but from only raw byte sequences. Our last experiments are conducted in section 7, where we discuss a calibration issue observed in our results, and how it can be remedied with minimal additional labeled data. We conclude in section 8 with suggestions for future work.

## 2 Related Work

In this work we restrict ourselves to information found in the PE header. A number of prior works have presented experiments which use only PE header information and obtained high accuracies, showing this to be a reasonable approach. One of the more thorough examinations of features from only the PE header was done by Shafiq et al. [53], in their PE-Miner project. They used a total of 189 features, 73 of which were binary features indicating the presence or absence of specific DLL library imports. They reported an overall Area Under the ROC Curve (AUC) of 0.991 for distinguishing between benign and malicious executables.

Schultz et al. [50] presented experiments using multiple feature types, including some using only PE header features. They used a list of DLL imports and function imports, obtaining 89.4% accuracy. Similar work considered only imported functions and obtained over 90% classification accuracy [30]. Elovici et al. [11] used imported functions as well as PE header fields, such as machine type and creation time. Using a Decision Tree as their classification model, their approach obtained 95.5% accuracy. Menahem et al. [32] used functions and fields from the PE header combined with other features.

In summary, previous work indicates that domain knowledge in the form of information gleaned from the PE header is effective at building classifiers that distinguish between benign and malicious executables. We therefore do likewise, by creating a baseline using only fields from the PE Header, as described in subsection 3.1.

Previous work has used domain knowledge from PE headers to train neural networks for malware detection [51]. Attempts have been made to use neural networks on top of byte n-gram features [11, 27, 36, 64]. In such an effort, the presence or frequency of specific n-grams is used as input to a neural network, and its output is an indication of whether the suspect software is benign or malignant. To train such a neural network, samples of benign and malicious software are used. While using neural nets on top of n-grams remains a domain-agnostic approach, it also inherits the shortcomings of byte n-grams, which include a partial loss of character sequence information, and a computational burden once $n$ grows beyond, say, $n = 4$. Similar work with neural nets for malware detection has been done with other feature types, such as op-code n-grams [35]. However, all of these approaches still require the initial feature specification and feature selection. In our approach, detailed in section 4, we look at doing the least possible amount of pre-processing, and allowing the neural network to build all of its needed representations, i.e. construct its own features, from the raw data.

We are interested in having a neural network learn its own feature representation from raw data due to successes in other domains, where this approach has lead to reduced domain knowledge and increases in accuracy simultaneously. For image classification Krizhevsky et al. [26] provide the first significant use, where they reduced the top-5 error rate on the ImageNet LSVRC-2010 contest from 25.7% using classical image processing feature extractors down to 17.0%. Recent work using related techniques has reduced that error rate to 5.71% [21]. For text classification Zhang et al. [65] used convolution networks on a variety of datasets, and found them to out-perform classic approaches on the four largest datasets, which where also the most challenging datasets evaluated. Similar progress has been made in machine translation, with recent results from Google approaching average human quality [58]. Our goal is to work towards a domain knowledge-free system for classifying malware. If successful, such a system could potentially be used for other similar domains (such as iPhone, Android, or Linux malware) with minimal work, providing additional utility through flexibility.

---

[2] We assume that the reader has a basic knowledge of neural networks. For those who seek a firmer background in neural networks and deep learning, we suggest Goodfellow et al. [17].

In subsection 4.2, we explore the potential for Recurrent Neural Networks, which process sequences of feature vectors, for this malware detection problem. Hidden Markov Models (HMMs) are another sequence based model that has been applied to the tasks of malware detection and family classification. Wong and Stamp [57] used HMMs over assembly op-codes to model specific malware families. Shafiq et al. [52] looked at first order HMMs over byte n-grams to model benign files, and this work is the most direct counterpart to our RNN approach. For detecting malware embedded into benign executables, they report a true positive rate of 84.9% and false positive rate of 16.7%. Their approach requires no domain knowledge, but is limited by the computational tractability of HMMs: to model $m$'th-order dependencies between $S$ states, a HMM requires $O(S^{m+1})$ parameters. Our RNN approach is able to learn higher order dependencies over the whole sequence. HMMs also lack the ability to create hierarchies of features, a demonstrated benefit of neural networks [46].

Some previous work has been done to apply RNNs to problems related to malware analysis. Pascanu et al. [40] applied RNNs to the task of malware detection, but used high level features hand-engineered by humans and collected through dynamic analysis. We instead seek to have the RNN learn any needed features for us, and we are restricting ourselves to static analysis. Shin et al. [54] used RNNs to identify function boundaries through partial disassembly of binaries. Their work uses byte inputs in a similar way to our approach, but for a different task. The problem tackled by Shin et al. is also simpler, requiring only a single hidden layer of 32 neurons to accomplish the task. As far as we are aware, no other work has yet considered using neural networks for malware detection from raw byte sequences.

# 3 Baseline Approaches

We now consider two baseline approaches for classification using information only from the PE header. Our first and primary baseline will extract these features using an existing library designed to work with the malformed headers present in malicious binaries. We will then discuss what of this domain knowledge is needed to extract the bytes where our features are stored. And finally we will present a simple approach using current domain knowledge-free methods on these bytes. These bytes will also be used as the only inputs to our neural networks described in section 4.

## 3.1 Domain Knowledge Approach

For our primary baseline we built a system similar to PE-Miner, restricted to only the first three headers of a PE file. These three headers almost always occur within PE executables and are easy to extract. By using these headers in particular, and not fully replicating the PE-Miner approach, we can ensure that all methods we evaluate here have equivalent information available for learning from. We use the PortEX library [20] to extract 115 features, of which 112 are numerical (such as the pointer to the Import table) and 3 are categorical (such as the intended runtime architecture), from the header of an executable. The PortEX library is specifically designed to work with real malware headers, which do not always conform to the official specification [10]. We use a subset of the features PE-Miner used so that all of our approaches tested are obtaining their features from the same fixed byte range of a binary, making the theoretical information available to all approaches equivalent. In early development we tested multiple different libraries for parsing the PE headers of malware, and did not always obtain consistent results on malware data. We settled on use of PortEX as it ran successfully on the most malicious binaries in a small test and was easy to use.

For our classifier, we tested the Random Forests [4] and the Extra Random Trees [16] algorithms. Both of these tree-based methods support categorical and numerical features. A tree based approach is favorable for this feature set since they are scale invariant. Our extracted features naturally have very different scales: some are binary 0/1 numeric values, and others contain an offset or field size, which could be a large integer. We used 100 trees for each model. Boosted decision trees were also tested, but did not perform as well and took longer to train, so their results are not included. We also tested several non-tree-based algorithms, such as linear and kernel SVMs, but they did not perform as well as any tree based approach for this feature set.

*Domain Knowledge Needed for Other Approaches*

While reviewing our domain knowledge approach of using header fields for feature vectors, we take a moment to discuss what subset of this knowledge is needed to perform our other two approaches. Indeed, this subset is extremely limited: the only necessary information is the length of the header (in bytes) and the PE header offset location.

According to the PE specification [33], the MS-DOS header is always the first 64 bytes of an executable binary. The only domain knowledge our other approaches need is the value of the PE header offset at byte 0x3C. The COFF and Optional PE header information is contained in the next 248 to 264 bytes following this offset, depending on whether the program binary is for 32 or 64 bit machines. We can always take the 264 bytes for safety, and concatenate it with our MS-DOS header, giving us 328 bytes to extract and use for our *almost* domain knowledge free approaches. In the rare cases that a file ends before the next 264 bytes can be obtained, we simply pad with zeros. In total this takes only a handful lines of code to extract, and is considerably smaller and less effort than the libraries that exist for parsing the PE header.

All of the information used by our domain knowledge approach is fully contained in these 328 bytes. Our other approaches must learn to extract that information from the raw bytes, which is a lower form of representation. Successfully extracting that information and obtaining accuracies comparable to the tree based domain knowledge approach would be an advantage in building reliable classifiers, avoiding the difficulties of extracting the features in practice.

While many previous papers have used the contents of the import table [11, 30, 32, 50, 53] we decided to exclude the import tables from our work. One of the main objectives of this paper is to minimize the amount of domain knowledge needed, and including such features would have interfered with that goal.

3.2 Byte N-Gramming Approach

Following the approach of Raff et al. [45], we use n-grams as features such that a feature vector has binary zero/one values if an n-gram is absent or present respectively. From these feature vectors, a model is created using an Elastic-Net regularized Logistic Regression classifier [66], the objective of which is given in Equation (1). This technique was chosen because it performs automatic feature selection as part of the regularization, allowing us to consider many different feature set sizes in a single, computationally efficient process.

$$f(w) = \frac{1}{2}||w||_1 + \frac{1}{4}||w||_2^2 + C\sum_{i=1}^{N} \log(1 + \exp(-y \cdot w^\mathsf{T} x_i)) \qquad (1)$$

The value $C$ in the loss function is the regularization parameter. Larger values of $C$ decrease the strength of the regularization; as $C \to \infty$, (1) approaches the behavior of standard Logistic Regression. Smaller values of $C$ reduce the effective degrees of freedom of the model, and force coefficients of $w$ to become zero.

Instead of n-gramming the whole file, our n-grams are extracted from only the aforementioned byte region associated with the PE headers. This approach of n-gramming only the header information is similar to the approach of Stolfo et al. [56], which n-grams only the beginning and ending bytes of a file, under the assumption that malware would most likely be found in those regions. We include the byte n-gram approach as a baseline comparison for an approach using minimal domain knowledge. We did testing of multiple different values of $n$ and used the model with the best performance in evaluation. Further details of the n-gram results can be found in Appendix A.

4 Neural Network Approach

We investigate two different types of neural networks for this task. Neural networks in general have made significant progress in speech [19] and image classification [26], where they are given low level features and construct their own higher level representations from just the examples given. This behavior is particularly desirable for our use case, since the format and behavior of a PE file is complicated. A malicious author may intentionally violate the rules of the format, making domain knowledge approaches fragile. A neural network could learn the way PE files are used in practice to increase performance and reduce demand on the developer to write additional code to process corrupted headers or handle corner cases. An added benefit is that a network will not have any kind of technical error on new and novel inputs, where a hand developed feature extraction process might. All of our networks were implemented in the Keras framework [6]. For readers more comfortable with neural network background, we invite them to skip to section 4.2 for details on our attention mechanism. The Keras code provides a more concise description of our architecture and can be seen in Appendix B.

4.1 Fully Connected Neural Networks

Our potential features consist of 328 ordered bytes to feed to our model. This limit makes it possible to use a fixed sized feature representation, meaning that we can use a Fully Connected (FC) Neural Network.[3] With the goal of working with minimal domain knowledge, we give these bytes directly to our model. Each byte is converted to a feature vector using a word2vec style embedding layer [34].[4] The embedding layer maps each byte value to a unique feature vector in $\mathbb{R}^B$, where $B$ is a hyper-parameter which defines the dimension of the embedding space. For our work we use $B = 16$, and found no performance difference when testing larger dimension embeddings. After each byte is embedded, we concatenate the 328 feature vectors into one large feature vector in $\mathbb{R}^{5248}$ to use as the input to the following layer. Note that the embedding layer is optimized by the learning process.

---

[3] A FC Neural Network is one in which the connections between nodes are acyclic.

For the architecture of our network, we used four fully connected hidden layers. During testing, we found that the each regularization technique we added to our network (as described below) increased the highest generalization accuracy achieved. While this is not a new phenomena, we note that the number of regularization methods we had to apply is somewhat unusual in comparison to modern approaches in other domains.

We used Dropout [55] for regularization with 50% probability on the hidden layers and 20% probability on the embedding space. Batch normalization [23] was applied after each hidden layer, which improves convergence and also has a regularizing effect. This was followed by the Exponential Linear Unit (ELU) as our activation function [7]. ELUs are one of many extensions to the successful ReLU [37] activation function, and attempts to address some weaknesses of the ReLU while retaining the faster convergence that it provides. We also applied the recent DeCov regularization [8] to the activations of the final hidden layer, using default parameters suggested by Cogswell et al. The DeCov regularization penalizes correlation between the activation values of the hidden layer, and its application affects all prior layers in the network. Last, we included a standard weight decay ($L_2$ norm) penalty of $10^{-4}$ for each hidden layer, with the first hidden layer also having an $L_1$ norm penalty. The $L_1$ norm has been found to be effective when dealing with high dimensional problems and induces sparsity, and was included since our input to the first layer has 5248 dimensions.

### 4.2 Recurrent Neural Networks

While we have a fixed length representation of bytes to feed to our neural network which enables us to use feed forward networks, we also experiment with Recurrent Neural Networks. RNNs are networks that process a sequence of input vectors, and carry a hidden state representation from one sequence to the next. Specifically, we use the Long Short-Term Memory (LSTM) model of Hochreiter and Schmidhuber [22] with forget gates [15]. These models have considerable additional complexity and computational restrictions compared to the feed forward architecture described in subsection 4.1, and as a result are not as practical for the task at hand.[5] Our interest is in seeing if RNNs are capable of learning the higher level features needed to make a benign or malicious classification from the low level features and sequential nature combined. This is important since RNNs can handle variable length sequences, making them good candidate models for future work in domain knowledge free classification of executables.

In our work we are restricting ourselves to just the header, making the sequence of bytes to examine much smaller than it may otherwise be. In the case of learning an LSTM, this is a necessary reduction to make the problem computationally tractable. Processing all bytes of a typical executable file one by one would result in millions of time steps. Learning a sequence of such length is far beyond the results of any work known to the authors. (Graves [18] trained generative models of text on up to 10,000 characters of history by using truncated back-propagation, however generative models benefit from an error signal that can be propagated at every step; this approach is not applicable to a classification task as opposed to a generative one.) For computational efficiency, we will use an embedding layer to process the input bytes, which are then fed into the RNN at each step.

We used a model with three LSTM layers, with an attention mechanism (explained shortly) before the classification. An example of the architecture we used is in Figure 1. Similar to our feed forward networks, we found additional regularization helpful in improving performance. However the recurrent nature of these models can change the nature of the impact a regularization has. The DeCov regularization we used in subsection 4.1 inhibited learning for this problem in particular, so was not included.[6] We used a Dropout probability of 50% in the method suggested by Gal [14] for regularization. For training we performed gradient clipping as presented by Pascanu et al. [39], clipping all norms to 1.0, and we performed gradient updates using the Adam update scheme [24].

---

[4] We also tested using a one-hot bit encoding of each byte, but found no difference in accuracy. We preferred the embedding layer since it is built into the Keras framework.

[5] It took about 11 days to train each LSTM model, compared to less than an hour for the Fully Connected network, all using a powerful Titan X GPU. All other methods presented in this work took 2 minutes or less to train on a 10-core workstation.

[6] We have successfully used the DeCov regularization with RNNs for other tasks, so we do not believe this is an intrinsic incompatibility.

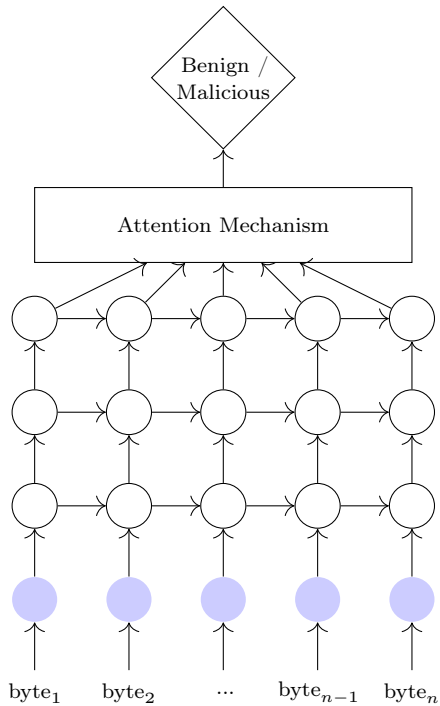

*Fig. 1.* Attention LSTM architecture used. Blue, shaded circles indicate embedding layers, white circles indicate LSTM layers. A 3-layer LSTM processes each byte in sequence, and the activations of all time steps are merged into the attention mechanism. The result from the attention mechanism is then used to make a classification decision.

## Attention Mechanism

Attention mechanisms have become an active area of research in Recurrent Neural Networks. The basic motivation is to improve performance by having the model increase or decrease the consideration and contribution for certain portions of the input, and have the added benefit of often being an interpretable component of the larger model. Here we briefly describe the attention mechanism used in our LSTMs, which is based on the approaches used by Yao et al. [59] and Rocktäschel et al. [49]. Both of these works use multiple sequences as inputs to the attention mechanism, and is a common scenario for their application. The attention is used to determine which items in one sequence are relevant to a specific item in the other.

In our case we have only one input sequence, the hidden states of the network. While our attention mechanism uses the same general components, the method of its application can be seen as a simplification for the case of sequence classification. We use the same basic building blocks but focus one part of the attention to the input sequence, and the other part to a summary of the input as a whole. In this way we contribute a method to attend to a subset of the input sequence, and give the details below.

The output of our attention mechanism is a weighted average of the hidden states (2).

$$\text{Attention Output} = \sum_{i=1}^{T} \alpha_i \boldsymbol{h}_i \tag{2}$$

Here $\boldsymbol{h}_i$ is the hidden activation of the last LSTM layer at step $i$. The summation weights $\alpha_i$ are also a part of the model, and $\sum_{i=1}^{T} \alpha_i = 1$. Given $T$ steps of the RNN, we compute the average hidden state activation over all time steps as $\overline{\boldsymbol{h}} = \frac{1}{T} \sum_{i=1}^{T} \boldsymbol{h}_i$. The vector $\overline{\boldsymbol{h}}$ provides information about the entire sequence, and each $\boldsymbol{h}_i$ provides information about one step in the sequence. These are combined in a small 1-layer network (that is a part of the larger RNN) as show in (3).

$$\widetilde{\alpha}_i = \boldsymbol{v}^\mathsf{T} \tanh\left(\boldsymbol{W}_0^\mathsf{T} \boldsymbol{h}_i + \boldsymbol{W}_1^\mathsf{T} \overline{\boldsymbol{h}} + \boldsymbol{b}\right) \tag{3}$$

The matrices $\boldsymbol{W}_0$ and $\boldsymbol{W}_1$ are for the attention mechanism's hidden layer, focusing on the local and global hidden state activations respectively. A single vector $\boldsymbol{v}$ is used to obtained the unscaled importance at each time step and $\boldsymbol{b}$ is the bias term. Finally, the softmax function is applied to produce a weighted importance over each time step that sums to one.

$$\alpha_i = \frac{\exp(\widetilde{\alpha}_i)}{\sum_{j=1}^{T} \exp(\widetilde{\alpha}_j)} \tag{4}$$

For our implementation, we include batch normalization and dropout for the attention mechanism. Experiments using multi-layer attention networks inhibited learning and did not perform as well as the one layer approach outlined above.

## 5 Experimental Methodology and Results

We have now described a total of five models we wish to evaluate. Extra Random Trees (ET) and Random Forests (RF), which will be our domain knowledge baselines trained on PE header features parsed out into their logical components. Logistic Regression on byte n-grams (LR) from the PE header area, which would be the classical method to do a domain knowledge free approach to this task. These methods were implemented using the JSAT library [44]. Last would be our two neural networks (FC and LSTM), which are also trained from only the raw bytes of the PE header. The neural networks were implemented with Keras [6].

To evaluate the results of a malware classifier most prior work uses the metrics of accuracy or Area Under the Curve (AUC), which are defined respectively as the percentage of executables correctly marked as benign or malicious, and the integral of the receiver operating characteristic (ROC) curve. These quantities are evaluated using either cross validation (CV) or on a held out testing set. The AUC can also be interpreted as the quality of the ranking a classifier produces of all data in the testing set. When ordered perfectly an AUC score of 1.0 is obtained. We report both metrics in this work, but use the Balanced Accuracy [5] where correctly labeling all benign files accounts for half of the final accuracy score and correctly labeling all malware account for the remaining half. This essentially re-scales the two classes to have equal weight.

Table 1. Breakdown of the number of malicious and benign training and testing examples in each data group, along with the sources from which they were collected. "Misc." comprises `portablefreeware.com`, Cygwin and MinGW.

|  | training | | testing | |
|---|---|---|---|---|
| Group A | malicious | benign | malicious | benign |
| Virus Share | 175,875 | — | 43,967 | — |
| Open Malware | — | — | 81,733 | — |
| MS Windows | — | 268,236 | — | 21,854 |
| Misc. | — | 1,195 | — | — |
| *total* | *175,875* | *269,431* | *125,700* | *21,854* |
| Group B | | | | |
| Industry Partner | 200,000 | 200,000 | 40,000 | 37,349 |
| *total* | *200,000* | *200,000* | *40,000* | *37,349* |

For training and evaluating these models we use the same data used in Raff et al. [45]. This data is divided into two groups and three testing sets, and the data sources are summarized in Table 1. Group A represents an approach to collecting data that others have used, namely, using malware from Virus Share [48] and binaries from Microsoft Windows to represent the benign set. An additional malware-only test set collected from Open Malware [43] is included as well. Group B is data provided by an industry partner, sampled from a larger corpus of benign and malicious executables. Group B is supposed to better represent the common types of files found on client machines.

We use the two different datasets to better evaluate how well our models generalize to new data. The approach used to assemble the Group A benign data is used by most prior works, and can lead to significant over-fitting on properties unique to binaries from Microsoft. This often fails to generalize to new data, as demonstrated by Raff et al. The Group B data appears to better represent common benign and malicious executables since the models trained on Group B maintain predictive ability on the Group A data, but the converse does not hold. We observe the same behavior using our domain knowledge features, indicating that even the limited information from the PE Header fields is enough to over-fit. This can be seen in Table 2, where we trained the domain knowledge approaches on the Group A training data. For this reason we do not further consider training on Group A data for evaluation in this work. We instead use the Group B training data for model construction, and consider the generalization to the Group B test set, as well as the Group A and Open Malware test sets.

Table 2. Performance of Random Forest and Extra Trees using header-based features. Models trained on only the Group A training data.

|  | Extra Trees | | Random Forest | |
|---|---|---|---|---|
|  | Accuracy | AUC | Accuracy | AUC |
| Group A Test | 99.1% | 0.999 | 99.5% | 0.999 |
| Group B Test | 69.9% | 0.723 | 71.5% | 0.728 |
| Open Malware | 99.9% | — | 95.5% | — |

We can see clearly that the models over-fit to the Group A data, even when limiting ourselves to only a subset of the PE Header. The PE-Miner project also used samples from their virology lab, but the ratio of files from Microsoft installs versus their lab is not mentioned. They report an AUC of 0.991, which is higher than what we get for our data trained with Group B, but similar to our 0.999 trained with Group A. This further leads us to suspect that the original PE-Miner results suffer from over-fitting. However, our suspicions notwithstanding, the test accuracies obtained when training on Group B (as presented in the next section) indicate the general approach of PE-Miner is sound.

Thus for our results all models are trained on only the Group B training data, and are evaluated against Group A, Group B, and Open Malware test data. For each method we have selected the parameters that resulted in the greatest performance. For n-grams we found 3-grams to perform best across all test sets, and more details can be found in Appendix A. The decision tree-based approaches were robust to parameter changes, with no noticeable difference in scores once 100 or more trees were used.

As neural networks are the primary interest of this work, we take a moment to give greater detail about their parameterization. We used three LSTM layers and four Fully Connected layers in our models, and tested networks with hidden state sizes of 128, 256, 512, 1024, and 2048 neurons. Due to computational constraints, we trained each parameterization for only 5 epochs, and then selected the best performing ones to train for additional epochs. Based on this we settled on a hid-den state size of 256 units for both architectures. We did not observe significant performance changes when increasing the number of neurons, and did some trial tests of larger networks and networks with more layers. Smaller networks performed slightly better while being more tractable. We suspect that the limited amount of data presented to the networks (only the first 328 bytes) compared to other recent works in image and audio processing meant that larger networks simply had too much extra capacity and flexibility, enabling overfitting. This suspicion is supported by the continued performance improvements we obtained when adding additional regularization techniques to our training process. Our final models were trained for 35 epochs, and the model from the best performing epoch was selected for evaluation.

5.1 Model Results

The final test set scores for our methods are presented in Table 3. We can see that the Fully Connected network is an improvement over our domain knowledge approach presented in subsection 3.1 in every metric. The network achieves higher accuracies by 3 to 4 percentage points and AUC improved by 0.053 for the Group B data. The Attention LSTM model doesn't perform quite as well as the domain knowledge approach, but is still considerably better than the byte n-gram approach of subsection 3.2. While the Fully Connected network is the more practical and appropriate model for this context, we believe the LSTM model is likely to be critical in future work since it can handle variable length sequences.

Below the ROC curves for all methods are presented for the Group A test set (Figure 2) and Group B test set (Figure 3). Due to the high AUC achieved by all methods on the Group A data, Figure 2 is zoomed in to the top left quadrant of the ROC curve. It is easy to see that the feed forward network dominates all other curves for Group A, and the 3-gram approach is dominated by all other curves. The LSTM approach and decision tree based approaches do better and worse in different regions of the ROC curve. On the Group B test set, the feed forward network almost dominates all other curves, with the Random Forest doing slightly better in a small region of the plot. These two methods generally did the best in most regions, followed by the other approaches overlapping each other in the plot.

Overall the feed forward network has performed better than all other methods, and the LSTM performed comparable to the domain knowledge approach at times, and was generally better than the byte n-gram approach. We note the importance of these results in demonstrating that current neural networks are able to learn from raw byte data. The binary content of the PE header, and of binary formats in general, is considerably different from the image, language, and signal processing tasks that neural networks have performed well on in recent years. All of the aforementioned tasks have a fairly continuous degree of spatial locality. Image and signal processing data can also be manipulated, transformed and corrupted while still retaining the original content and semantic meaning. This is applicable to language problems as well, though to a lesser degree, through the use of word replacement. Using corrupted data instances has proven useful in training neural networks and increasing model robustness, and is also an indication of the robustness of information in those formats. This is possible for all these tasks in part because the nature of the data and information represented is generally constant and has meaningful transitions when values change. For example, a photograph of an apple can be rotated, re-scaled, lightened, have noise added, etc. and still recognizably be a picture of an apple. The same can not be said of an executable, where the modification of even a few bits may lead to very different behavior.

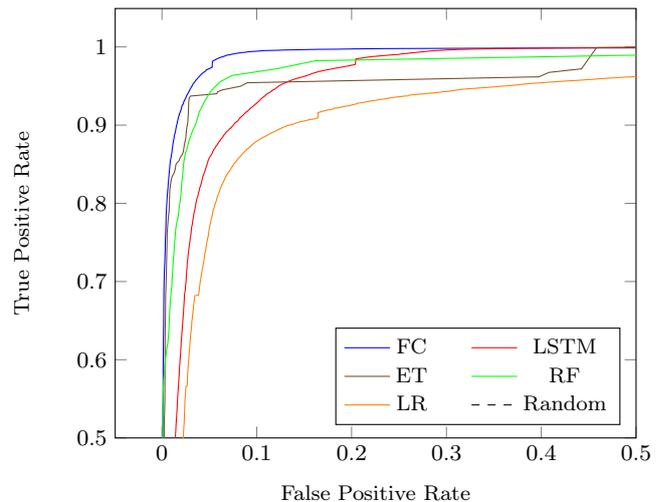

Fig. 2. ROC plot for all models on Group A test data. Figure zoomed in to the top left corner due to high AUC of all models.

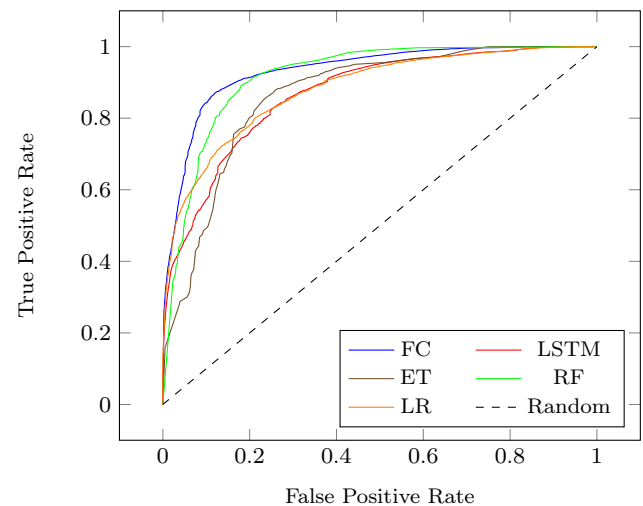

Fig. 3. ROC plot for all models on Group B test data.

Table 3. Performance of all methods on the test sets. Accuracy scores are balanced so classes have equal contribution.

|  | Group A Test | | Group B Test | | Open Malware |
|---|---|---|---|---|---|
|  | Accuracy (%) | AUC (%) | Accuracy (%) | AUC (%) | Accuracy (%) |
| Fully Connected | 90.8 | 97.7 | 83.7 | 91.4 | 89.9 |
| LSTM | 84.2 | 96.7 | 77.5 | 86.7 | 79.7 |
| Extra Tree | 86.4 | 97.2 | 80.7 | 86.1 | 85.5 |
| Random Forest | 78.9 | 96.8 | 82.3 | 91.2 | 64.4 |
| LR 3-grams | 71.2 | 91.4 | 77.8 | 87.3 | 61.5 |

In contrast, our binary data has some locality, but its nature is variable length, and abrupt discontinuities in locality are a constant impediment. Small perturbations in the byte values can have a dramatically different meaning, making it difficult to do any kind of augmented training. Even our limited PE Header data also has different modalities of information: one byte could be a bit-wise set of boolean variables, follow by a byte representing an integer offset, followed by another byte encoding a choice between one of many categorical options. Each has an intrinsically different nature that must be processed and extracted by the network in different ways. That the network is able to learn this classification problem, despite the raw data behavior differing considerably from image and signal tasks, is noteworthy. This provides evidence to the robustness of a neural network approach and that it is plausible to perform malware detection with reduced domain knowledge.

*Comparison with Whole File Byte N-Grams*

The results from Raff et al. [45], when byte n-gramming the whole executable, models trained on Group B obtained balanced accuracies of 87.3%, 94.5%, and 81.1% for the same Group A, B, and Open Malware tests sets. These scores indicate that the performance of only the header information is at some level comparable in generalization to those of byte n-grams over the whole file. The significant difference between the two approaches is with respect to the Group B test accuracy. We would generally expect the performance on Group B's test set to be higher, since the model was trained on data from the same source, as occurs with the byte n-gram approach. The tested approaches instead performed slightly worse on Group B test data compared to the other tests sets. These results are an indication that the information from the PE header can generalize especially well to new data when provided an appropriate training set. This also highlights the importance of testing on Group A and Open Malware to better evaluate the generalization of our model to different populations. While the improved generalization from the PE header is a desirable quality, the lesser performance on Group B is still an issue. Especially since we believe Group B's data better reflects the populations of both benign and malicious software. Future work may look to determine why the generalization characteristics of these approaches seem to differ.

## 6 Inferring Relative Feature Importance

In order to further confirm our results, we would like to understand what features have been constructed by our neural network model. Model interpretability is beneficial in diagnosing what a model has learned and confirming that it is picking up on reasonable features, as well as in detecting over-fitting. Understanding what a neural network has learned is a challenging task though, and in particular neural networks are often regarded as a "black box" model. There are many different ways to discuss and define what it means for a model to be interpretable [28], and a common approach for neural networks has been to visualize synthetic inputs that would maximally activate certain neurons (which could include output neurons) within a network [46, 47, 63]. Most of these approaches are motivated by the domain of image processing and unfortunately are not directly applicable for the malware domain. Especially since the synthetic inputs that are generated may not correspond to a realistic or plausible input.

To circumvent this issue, we look at the importance of features learned by one of our networks and compare it to the features used by our (more interpretable) domain knowledge approach. By establishing an overlap between the features found important for our neural networks and domain knowledge approach, we increase our confidence that the neural networks are learning reasonable features and not over-fitting to the problem.

One of the benefits of the tree based approaches is the ability to infer feature importance from the learned model. Unlike the lasso models used with n-gramming in subsection 3.2, these feature importance scores consider the non-linear interactions of the features as learned by the model. We use the Mean Decrease in Impurity (MDI) [3, 29] on the training data to find the top 10 most important features, which can be found in Table 4.

We used the Gini measure for tree construction, and so use it for measuring feature importance as well. Denoting $G(t)$ for the Gini impurity computed from all data that went through tree node $t$, we obtained the change in impurity $\Delta G(t)$ at node $t$ by $\Delta G(t) = G(t) - p_{t_L} G(t_L) - p_{t_R} G(t_R)$. Here $t_L$ and $t_R$ are used to denote the left and right child branches of the node $t$, and $p_{t_L}$ and $p_{t_R}$ the relative proportion of data that went to each child. If $t_s$ is the feature used at node $t$, $p(t)$ is the fraction of all data that reached node $t$, and $\mathbb{1}[x]$ is the indicator function, we can define the MDI for a feature $s$ of a tree model as (5).

$$\text{MDI}_s = \sum_{\forall t \in \text{Tree}} \mathbb{1}[t_s = s] \cdot p(t) \cdot \Delta G(t) \quad (5)$$

For an ensemble of trees, we average the MDI score from each tree to obtain a final score. The MDI is then maximized when a feature causes a large decreases in impurity and is used for larger portions of the dataset. This intuitively gives higher scores to features used more frequently that have a significant impact on the model's induction. We refer the reader to Louppe et al. [29] for a more thorough description of the MDI and its theoretical justification.

To aid in interpretation we show the relative importance (RI), where a features importance score is divided by the maximum importance value (i.e. $\text{RI}_s = \text{MDI}_s / \arg\max_j \text{MDI}_j$). We can see that the Extra Trees algorithm is focusing in on only a few feature types, with the relative importance quickly decaying.

We can also infer feature importance from the attention weights of the LSTM model. For each datum, the LSTM gives a weight to the hidden activation at each time step. We can interpret these weights as importance values for the byte presented at each aforementioned time step. By computing these attention values for each byte, and averaging over the training data, we obtain a method of ranking the importance of each byte as determined by our LSTM model. The relative importances from our LSTM do not decay as quickly. Part of this can be explained by the fact that the LSTM is given 328 distinct bytes, where the tree models are given 115 higher level features they represent. Given that the LSTM observes features byte by byte, and that the Extra Tree's domain knowledge features can be mapped to multiple byte ranges, it is difficult to do a direct top-10 comparison. Instead we group the top 70 bytes by value, as indicated by the attention weights. Comparing the bytes found to be used by the LSTM to the byte ranges of the features selected in Table 4 we discover multiple overlaps between what the LSTM selected and what the Extra Trees algorithm chose as the most important features.

Table 4. Most important features, as determined by the Extra Trees algorithm. Relative Importance (RI) is in the last column, with 1.0 for the most important feature. Byte ranges expressed as [min, max) when the feature is multiple bytes in length. For features that may have a different location for 32 and 64-bit binaries, the 32-bit location is specified first, followed by the 64-bit location.

| Header Field | Byte Location | RI |
|---|---|---|
| IMAGE_FILE_DLL | 87th, 3rd MSB | 1.00 |
| Certificate Table's size | [220, 224)/[236, 240) | 0.42 |
| TERMINAL SERVER AWARE | 159th. MSB | 0.34 |
| Export Table's size | [188, 192)/[204, 208) | 0.33 |
| CLR Runtime Header's size | [300, 304)/[316, 320) | 0.23 |
| Subsystem field | [156, 158) | 0.21 |
| Debug table's size | [236, 240)/[252, 256) | 0.18 |
| CLR Runtime Header's offset | [296. 300)/[312, 316) | 0.15 |
| Import Table's size | [196, 200)/[212, 216) | 0.12 |
| NO SEH | 158th, MSB | 0.09 |

The IMAGE_FILE_DLL (byte 87), TERMINAL SERVER AWARE (byte 159), and NO SEH (byte 158) byte fields were all selected. These three fields are single bit values within a larger byte sequence. The LSTM, because of the embedding layer, cannot directly observe the raw bit values. This is a positive indication the LSTM is able to learn finer grained meaning embedded in the byte information presented.

The whole byte range of the CLR Runtime Header's offset and size for 64 bit values (byte range [312, 320)) was selected. The two most significant bytes of the Certificate Table's size for 32-bit executables were chosen and the most significant byte for 64-bit executables. The most significant byte of the Subsystem field and 32-bit Debug table size were selected. The three most significant bytes were selected for the Import Table's size in 32-bit executables. The LSTM seems to often select only the most significant bytes of some multi-byte fields. This may indicate that the field's importance is in whether or not a large value was given, allowing the LSTM to avoid focusing on the lower order bytes. We also note that only the first and last bytes of the MS-DOS header were in the top 70 bytes by importance, indicating that the MS-DOS header did not play a major role in the decision process. This is corroborated by the MDI of those features which is near zero in terms of relative importance.

Overall the LSTM's attention placed more weight on bytes associated with nine of the top ten features chosen by our domain knowledge approach. The correspondence between information is not necessarily perfect. The Extra Tree features are ranked by importance, but some may or may not be used when a binary is 64-bit or 32-bit. The LSTM feature importance is derived by averaging the attention weights across all the training data samples, so we again lack this information in the aggregate case. Despite these interpretation issues, there appears to be meaningful overlap between what features our LSTM network are learning to use, without any domain knowledge, and what the Extra Trees algorithm selected when given the domain knowledge features. This is supports the validity of our approach, and our claim that neural networks can learn higher level features from raw byte information.

We would expect that our Fully Connected network has learned similar features, and likely learned them better since it achieves higher accuracies. While it would be desirable to affirm this case as well, our Fully Connected network lacks the same direct, intuitive, and interpretable feedback that our Attention mechanism provides. Future work may look into how to better infer feature importance for the malware domain for various other types of neural architectures.

## 7 Network Calibration

We have shown that our neural network method surpasses the accuracy of a more conventional approach to the PE header, and provided evidence that it learns similar concepts. We note though that all methods have shown evidence of a calibration issue, where the accuracies obtained are lower than one would expect given their AUCs. This unusual characteristic is particularly noticeable for our neural network approach when we evaluate one the test set after each epoch, which can be seen in Figure 4. Evaluating the test set performance after each training epoch, we see the AUC scores are stable and reasonably high, often reaching 0.91 for the B test set and 0.98 for the A test set. The Group B test set accuracy has some fluctuation, as does the Group A test set accuracy. The Open Malware test set, which contains only malware, has wide swings in accuracy — and no AUC score since it contains only one class. When testing multiple runs of the same architecture, we consistently obtained this behavior and consistently achieved similar peak accuracies and AUC scores. This behavior occurred for our Recurrent network as well, and can be seen in Figure 5. Based on the AUC performance one can infer that it takes multiple epochs for the LSTM to start learning the problem, where our Fully Connected network reaches high AUC scores after just one epoch through the data.

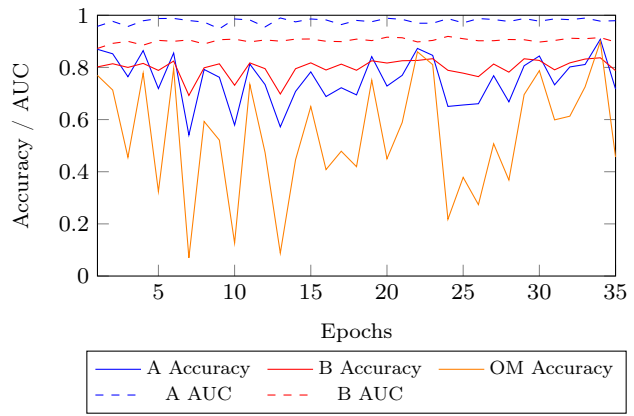

*Fig. 4.* Test set balanced accuracy and AUC scores for Fully Connected model. AUC shown as dashed lines, balanced accuracy as solid.

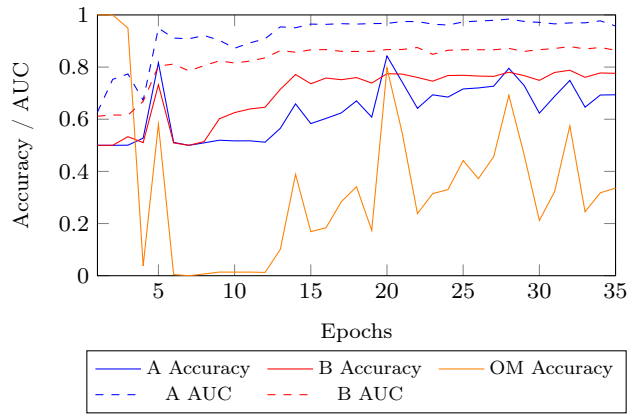

*Fig. 5.* Test set balanced accuracy and AUC scores for LSTM attention model. AUC shown as dashed lines, balanced accuracy as solid.

Considering that our model was trained on Group B data, it is important to note the somewhat surprising behavior that the AUC on Group B test data is *worse* than the AUC on Group A test data. The normal expectation would be that our model's performance should be highest on data most similar to the training distribution, Group B, and hopefully generalize (with a minimal degradation in accuracy) to other distributions, such as Group A and Open Malware. However, we contend that the observed behavior is a positive result for our approach. The Group A data is an objectively simpler distribution than Group B, with most if not all the benign files coming from Microsoft Windows. This particular population should be easier to separate from malware for a model that is generalizing well, and so it is reasonable to get a higher AUC.

Next is the issue of performance discrepancy between Accuracy and AUC. It is reasonable to expect that models achieving a high or low score in one metric should obtain a similarly high or low score in the other. Indeed it has been shown that the error rate and AUC of a classifier are directly related, but they can also obtain considerably different values depending on the data distribution used and the objective function of the model [9]. In our data these scores differ considerably, with especially large fluctuations for the Open Malware data. This unusual scenario can be caused by the fact that the AUC metric is a function of the model's ranking of data points, not by its ability to make a decision about the classification of the data. Two examples of how it is possible to have large discrepancies between AUC and accuracy are presented in Figure 6. Our results match the scenario of Figure 6a, where the best possible accuracy would be 90%, but a poor calibration in the decision threshold results in only 60% accuracy. Regardless of where the decision is made to change from circles to crosses, the AUC is very high at 0.96. Similarly, Figure 6b shows how a relatively high accuracy could be obtained despite having a low AUC of only 0.64.

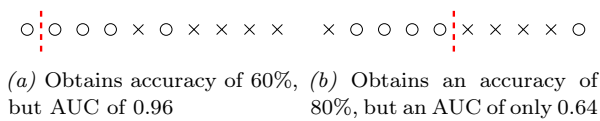

*(a)* Obtains accuracy of 60%, but AUC of 0.96

*(b)* Obtains an accuracy of 80%, but an AUC of only 0.64

*Fig. 6.* Example showing how the AUC can be larger or smaller than the accuracy. Two classes represented with circles and crosses. Relative ordering indicates classification ranking, with the dashed line showing the decision point.

The especially large swings in accuracy for the Open Malware data indicate that the model is regularly changing the crossover point between benign and malicious for that test set. Since the Open Malware data has only malicious examples, it drops below 50% accuracy when the decision point switches to preferring a label of benign, and dramatically increases when preferring a label of malicious. The Group A data, having members of both classes, has less dramatic yet pronounced swings as the decision threshold is adjusted. The Group B test set, being from the same distribution as the training data, has relatively minor changes. Ultimately, this would indicate that while our decision point is being learned well for the Group B data, it is not always well calibrated for different data distributions.

Given this behavior it is reasonable to be concerned with the classification accuracy when the model is applied to a new distribution. The consistently high AUC scores give us confidence that, if we can obtain limited labeled data for a new target distribution, the issue can be circumvented through the use of probability calibration techniques such as Platt's Scaling [42] and Isotonic Regression [62]. These methods use the prediction outputs as a small feature set to a new model. Platt's Scaling essentially performs Logistic Regression using the original model's prediction as the only feature. In this way these methods do not alter the decision surface, but instead shift the decision threshold. The issue of re-calibrating probabilities from limited data was examined by Niculescu-Mizil and Caruana [38], and they found that as few as 100 data points for calibration could significantly improve performance.

We demonstrate this possibility in Figure 7, where a small portion of the test set is used (with labels) to re-calibrate the probabilities to the Group A data.[7] For this test we selected each model from the epoch that had the worst accuracy on the test set. This evaluation is thus using a different set of weights for the neural network than what was used to get our results in section 5. When selecting the networks epochs that had the worse accuracy, the Group A test accuracies were 54.1% for the Fully Connected network and 57.5% for the LSTM network. By using just 20 calibration points both models have dramatically improved accuracies, reaching 93.3% and 75.1% for feed forward and LSTMs respectively. Our Fully Connected network performed better even after calibrating from just 10 labeled data points.

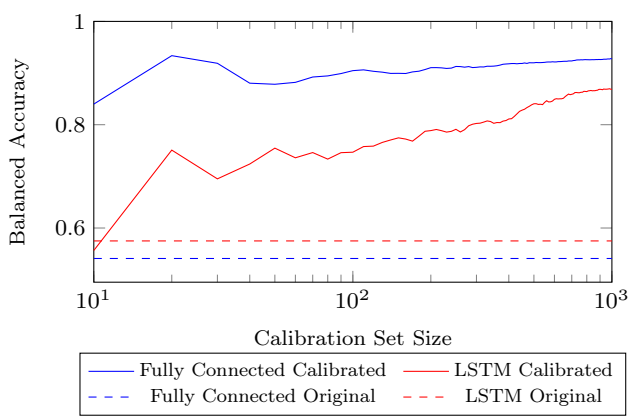

*Fig. 7.* Accuracy of neural network models trained on Group B data, and then calibrated with a small amount of the Group A test data and tested on the remainder of the Group A test data. Selected models from the training epoch that had the *worst* generalization performance to Group A to be calibrated.

---

[7] Platt calibration is normally done off the scalar product between the weight vector and the data vector, before the logistic function is applied. Due to code limitations, we performed calibration on the output probabilities (i.e. after the logistic function). This may make our results sub-optimal.

More advanced calibration could be done by retraining only the last layer of the neural networks, keeping all other weights the same. This is equivalent to using the penultimate activations of the network as a feature vector and training a new logistic regression model from those features. Re-training the final activation weights learns a new decision surface, though linear with respect to the already learned representation. This would require more labeled data than simple re-calibration schemes such as Platt's Scaling, but may further extend the utility of a pre-trained model. Prior work has shown it is possible to fine tune the weights of a neural network to new tasks [12, 60], and combined with our results on re-calibration indicate that this approach would also be viable.

While we have shown how to resolve the calibration issue with very limited labeled data, we have yet to ascertain why this calibration issue occurs. Both the Random Forest and n-gram models also have lower accuracies than one would expect for their AUCs that they achieve. For this reason we suspect this issue may some how be related to the feature source (the PE-header), and not the model or feature representation. The issue is more apparent for the neural network approaches simply because we plot the accuracy as a function of training epoch, which shows the fluctuations occur over training. Determining and better resolving the cause of this fluctuation is an open problem for future work.

## 8 Conclusions and Future Work

In this paper we have shown the potential for neural networks to learn from raw byte values, demonstrated by classifying executables as benign or malicious without any feature engineering or processing. Restricted to a manageable range of bytes from the PE Header, our networks are able to match and even surpass the performance of a domain knowledge approach that has equivalent information available. To wit, this is the first application of neural networks on raw byte sequence features for this task, and does not suffer the pitfalls of the domain knowledge approach. We have also shown our model's underlying representation is robust enough for re-calibrating models to new domains. This may be significant to larger corporations and organizations that may receive targeted or otherwise unique intrusion attempts.

In future work, we intend to look at increasing the amount of information available to a network and work toward processing the whole executable with minimal domain knowledge. Our re-calibration results may also be of interest in malware family classification. In the spirit of one-shot learning, a robust model may be better suited to recognizing newly discovered malware families after only a few examples. This could be valuable in determining how widespread a newly discovered virus type is, without having to manually construct signatures for the virus.

## Acknowledgment

We would like to thank the Laboratory for Physical Sciences (LPS), where this work was performed and Mark McLean for supporting it. We would like to thank the Nvidia Corporation for their assistance and for providing GPU compute nodes for this work. In particular, we would like to thank Jon Barker for his help in managing these resources for us. We would also like to thank Robert Brandon for reviewing drafts of this paper and helpful discussions.

**Appendix A    N-Gram Details**

In this section we provide further details on our work in using byte n-grams on just the PE header. We have seperated this section as it does not add to our goal of showing that neural networks can learn from raw byte data. However, we do wish to show that we attempted to maximize the performance of the byte n-gram approach to form as strong a baseline as possible.

For our implementation of this algorithm we used a modified version of the newGLMNET algorithm [61]. By using the warm-start strategy from Friedman et al. [13], we can efficiently train models for numerous values of $C$ at only incremental cost. This reduces the time for performing a parameter search and produces a regularization path, where we look at a property of the model (such as accuracy or number of non-zero weights) as a function of the regularization parameter.

Normally byte n-gramming of entire files is computationally expensive, especially when trying multiple configurations. In such a case it is generally beneficial to perform feature selection on the n-grams extracted due to computational constraints and to reduce the impact from the curse of dimensionality [1, 2]. This can be done by using information-gain or simply removing features that do not reach a minimum frequency [25, 41, 45]. Extracting the 328 bytes from our headers of interest significantly reduces the amount of data to process, increasing the flexibility of what we can experiment with when using n-grams. For example, there are more than 4 billion unique 6-grams in our training data of headers, and only 1,218 unique 6-grams that occur in at least 1% of the header files. The more rigid structure of the PE header also reduces counts for 2-grams: of the 65,532 seen, only 563 occur in at least 1% of the header files. Thus it is computationally feasible for us to evaluate n-grams $\forall n \in [2,6]$. We also evaluate the merging of all n-gram features into one larger feature set, where each file is n-grammed for each value of $n$ and concatenated into one large feature vector for each file. Using all n-grams in one feature set could potentially help avoid a mismatch between the underlying data, where a field may be stored in 1 to 4 contiguous bytes, and a fixed n-gram size.

In Table 5 we show the balanced accuracy and AUC on our test sets for groups A and B, as well as the recall on Open Malware data. Over all test n-gram sizes, 2 and 3-grams performed the best when considering the recall on Open Malware data, but had slightly worse generalization to the A and B test sets compared to large n-gram sizes. However, these n-gram models don't generalize as well as those that use domain knowledge, in the sense of poorer performance when applied to another dataset. Looking at the regularization path of 3-grams in Figure 9, we can also see evidence of over-fitting as the Group A test set performance at first increases, and then drops while the model is still learning and improving the performance on the Group B data. Similar behavior can also be seen from 2-grams in Figure 8. The plot also shows a significant 15 point gap between the 10-fold cross validation accuracy and the accuracy on B's test set. This is significant since the model was trained on Group B data. The model is thus failing to generalize in two ways: from training data to testing data of the same distribution (Group B training to Group B testing), and from the training distribution to a new distribution (Group B to Group A).

In figures Figure 10 and Figure 11, we can see this learning curve behaves worse for the larger n-gram sizes, where the Group A accuracies only decrease as the model learns to fit the Group B training data. This indicates an even high level of overfitting. This is congruent with the specificity increasing (and therefore generalizability decreasing) with n-gram sizes. These regularization paths provide further evidence that 2 and 3-grams are better than $\geq 4$ grams for PE headers.

*Table 5.* Performance of byte n-grams trained on the Group B training data using Elastic-Net regularized Logistic Regression.

|  | 2-grams | 3-grams | 4-grams | 5-grams | 6-grams | [2,6]-grams |
|---|---|---|---|---|---|---|
| A Accuracy | 72.4% | 71.2% | 71.1% | 70.9% | 71.3% | 71.5% |
| A AUC | 0.931 | 0.914 | 0.951 | 0.964 | 0.955 | 0.956 |
| B Accuracy | 76.6% | 77.8% | 78.9% | 79.8% | 79.4% | 79.9% |
| B AUC | 0.857 | 0.873 | 0.879 | 0.885 | 0.884 | 0.897 |
| OM Recall | 62.6% | 61.5% | 51.4% | 53.3% | 45.0% | 51.7% |

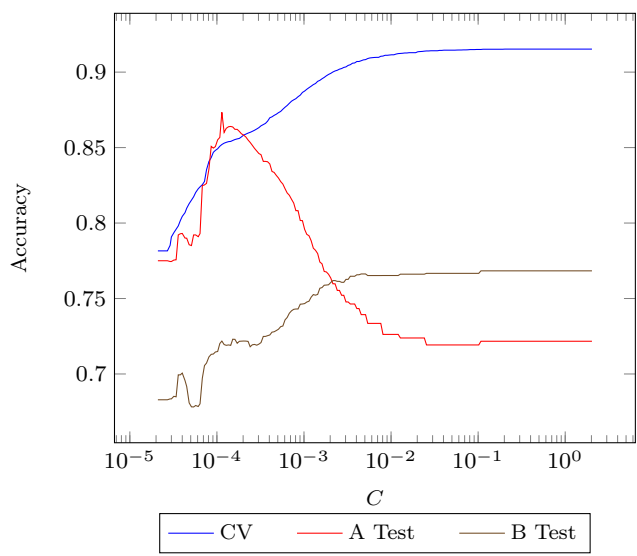

*Fig. 8.* Regularization path of 2-grams, with 10-fold cross validation and test set performance measured in weighted accuracy.

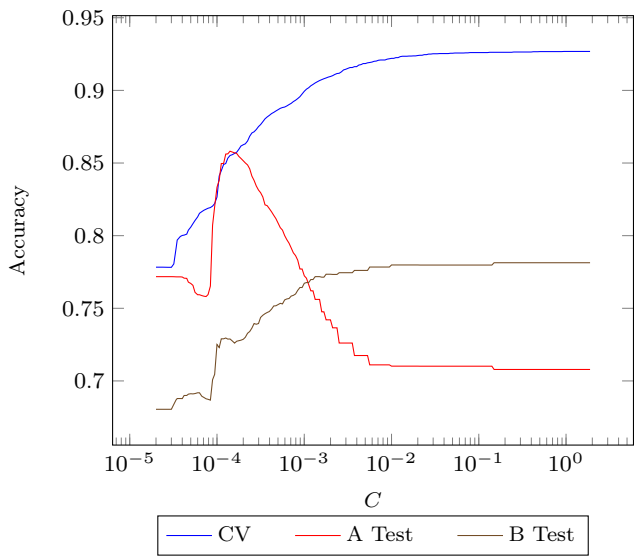

*Fig. 9.* Regularization path of 3-grams, with 10-fold cross validation and test set performance measured in balanced accuracy.

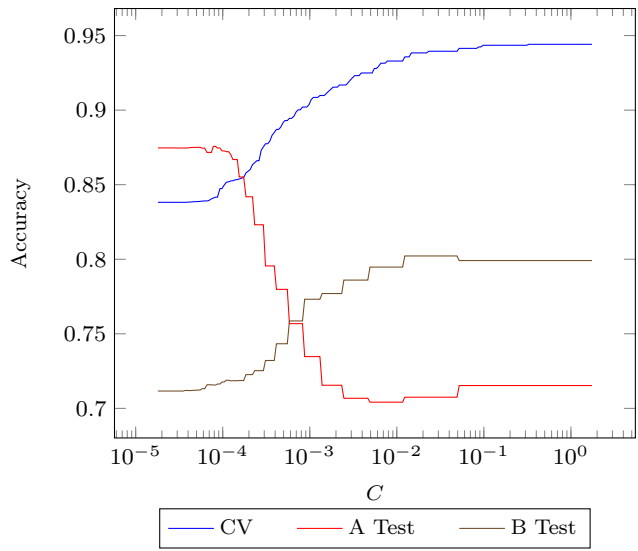

*Fig. 11.* Regularization path modeling using 2 through 6 grams, with 5-fold cross validation and test set performance measured in weighted accuracy.

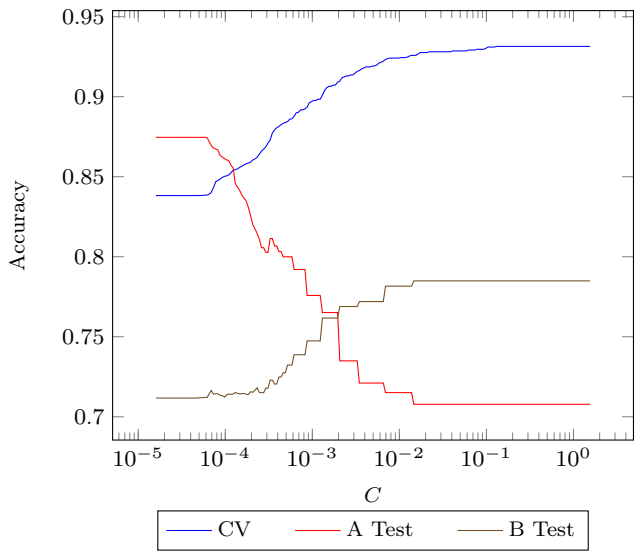

*Fig. 10.* Regularization path of 6-grams, with 10-fold cross validation and test set performance measured in weighted accuracy.

indicate that the 3-grams are more likely to generalize to new data than the 6-grams, which degrade to worse than 50% recall.

We also note again the importance of evaluating generalization of executables over multiple corpora, as otherwise we would have erroneously considered 5 and 6-grams to perform best in this scenario, since they have an almost 2 point advantage over 3-grams when tested on the Group B data. Yet when tested on different distributions (Group A and Open Malware), 5 and 6-grams perform considerable worse than 3-grams. This is important in the context of a malware classifier, as a deployed model may see data considerably different from what it was trained on. The Open Malware performance would

# Appendix B  Keras Model Definitions

For the specifics in how our neural network architectures are organized, we include partial Python code snippets of the model definitions. These do not include the definitions for custom layers, but are meant to provide more precise details of how our architecture is designed for those who are interested. All models were implemented using version 1.0.4 of the Keras library.

In both of the code snippets below, `maxlen = 328`, indicates the length of the byte sequence and `embed_size = 16` is the size of our embedding layer. The variable `main_input` represents the initial sequence of bytes given to the model. The Fully Connected network is defined in Listing 1. In Listing 2, `S` is the number of neurons in the LSTM layers.

```python
main_input = Input(shape=(maxlen,), dtype='int32',
    name='main_input')
emb = Embedding(256, embed_size, input_length=maxlen,
    dropout=0.2, W_regularizer=l2(1e-4))(main_input)

x = Flatten()(emb)

num_layers = 4
for i in range(num_layers):
    W_reg = l2(1e-4)
    if i == 0:
        W_reg = l1l2(1e-4)
    x = Dense(layer_size, activation='linear',
        W_regularizer=W_reg)(x)
    x = BatchNormalization()(x)
    x = ELU()(x)
    if i == num_layers-1:
        x = DeCovRegularization(0.1)(x)
    x = Dropout(0.5)(x)

loss_out = Dense(1, activation='sigmoid',
    name='loss_out')(x)

model = Model(input=[main_input], output=[loss_out])
optimizer = Adam(lr=0.001)
model.compile(optimizer, loss='binary_crossentropy')
```

*Listing 1.* Keras model definition for Fully Connected network

```python
main_input = Input(shape=(maxlen,), dtype='int32',
    name='main_input')
emb = Embedding(256, embed_size, input_length=maxlen,
    dropout=0.2, W_regularizer=l2(1e-4))(main_input)
hs = [] #hidden states from each LSTM layer stored here
hs.append(LSTM(S, dropout_W=0.5, dropout_U=0.5,
    W_regularizer=l2(1e-5), U_regularizer=l2(1e-5),
    return_sequences=True)(emb))
for l in range(1, num_layers):
    hs.append(LSTM(S, dropout_W=0.5, dropout_U=0.5,
        W_regularizer=l2(1e-5), U_regularizer=l2(1e-5),
        return_sequences=True)(hs[-1]))
local_states = merge(hs, mode='concat')
average_active = AverageAcrossTime()(local_states) #this
    produces $\overline{h}$
state_size = lstm_layer_size*num_layers
#Attention mechanism starts here
attn_cntx = merge([local_states,
    RepeatVector(maxlen)(average_active)],
    mode='concat')
attn_cntx = TimeDistributed(Dense(lstm_layer_size,
    activation='linear',
    W_regularizer=l2(1e-4)))(attn_cntx)
attn_cntx =
    TimeDistributed(BatchNormalization())(attn_cntx)
attn_cntx =
    TimeDistributed(Activation('tanh'))(attn_cntx)
attn_cntx = TimeDistributed(Dropout(0.5))(attn_cntx)
attn = TimeDistributed(Dense(1, activation='linear',
    W_regularizer=l2(1e-4)))(attn_cntx) # $\widetilde{\alpha}_i$
attn = Flatten()(attn)
attn = Activation('softmax')(attn) # $\alpha_i$
attn = Reshape((maxlen, 1))(attn)
attn = TileOut(state_size)(attn) #repeats value to make
    a specific shape
final_context = merge([attn, local_states], mode='mul')
final_context = SumAcrossTime()(final_context) # eq
    (2), $\sum_{i=1}^{T} \alpha_i \boldsymbol{h}_i$
final_context = Dense(state_size, activation='linear',
    W_regularizer=l2(1e-4))(final_context)
final_context = BatchNormalization()(final_context)
final_context = Activation('tanh')(final_context)
final_context = Dropout(0.5)(final_context)
loss_out = Dense(1, activation='sigmoid',
    name='loss_out')(final_context)
model = Model(input=[main_input], output=[loss_out])
optimizer = Adam(lr=0.001, clipnorm=1.0)
model.compile(optimizer, loss='binary_crossentropy')
```

*Listing 2.* Keras model definition for LSTM Attention Network